\pdfoutput=1

\documentclass[10pt,twocolumn,letterpaper]{article}

\usepackage[pagenumbers]{wacv} 

\usepackage{graphicx}
\usepackage{subcaption}  
\usepackage{amsmath}
\usepackage{amssymb}
\usepackage{booktabs}

\usepackage{multirow}
\usepackage{pifont}
\usepackage{dsfont}
\usepackage{algorithm}
\usepackage{algpseudocode}
\usepackage{enumitem}

\usepackage[accsupp]{axessibility} 

\usepackage[pagebackref,breaklinks,colorlinks]{hyperref}

\usepackage[capitalize]{cleveref}
\crefname{section}{Sec.}{Secs.}
\Crefname{section}{Section}{Sections}
\Crefname{table}{Table}{Tables}
\crefname{table}{Tab.}{Tabs.}

\def\wacvPaperID{13} 
\def\confName{WACV}
\def\confYear{2025}

\begin{document}

\title{Leveraging Satellite Image Time Series for Accurate Extreme Event Detection}

\author{Heng Fang \hspace{.9cm} Hossein Azizpour\\
KTH Royal Institute of Technology, Stockholm, Sweden\\
{\tt\small \{hfang, azizpour\}@kth.se}
}

\maketitle

\begin{abstract}
    
    Climate change is leading to an increase in extreme weather events, causing significant environmental damage and loss of life. Early detection of such events is essential for improving disaster response. In this work, we propose SITS-Extreme, a novel framework that leverages satellite image time series to detect extreme events by incorporating multiple pre-disaster observations. This approach effectively filters out irrelevant changes while isolating disaster-relevant signals, enabling more accurate detection. Extensive experiments on both real-world and synthetic datasets validate the effectiveness of SITS-Extreme, demonstrating substantial improvements over widely used strong bi-temporal baselines. Additionally, we examine the impact of incorporating more timesteps, analyze the contribution of key components in our framework, and evaluate its performance across different disaster types, offering valuable insights into its scalability and applicability for large-scale disaster monitoring.
\end{abstract}

\section{Introduction}
\label{sec:intro}

Climate change is a tremendous threat to sustainable life on planet Earth. One of its key adverse outcomes is the increasing prevalence of extreme weather events such as floods, storms, hurricanes, and forest fires.  According to CRED (Centre for Research on the Epidemiology of Disasters) ~\cite{cred2023}, in 2023 alone, we experienced about 400 natural hazard disasters that affected a population of more than 90 million and claimed over 85,000 human lives; besides the total economic damages which reached an estimate of 202.7 billion US dollars. These devastating numbers and trends beg for actions towards prevention, mitigation, preparation, and most imminently, \textit{response}. An effective disaster response against such extreme events primarily requires early detection of their onset. In this work, we aim to use satellite image time series to detect the onset of an extreme event.

\begin{figure*}[h]
\begin{center}
\includegraphics[width=0.9\linewidth]{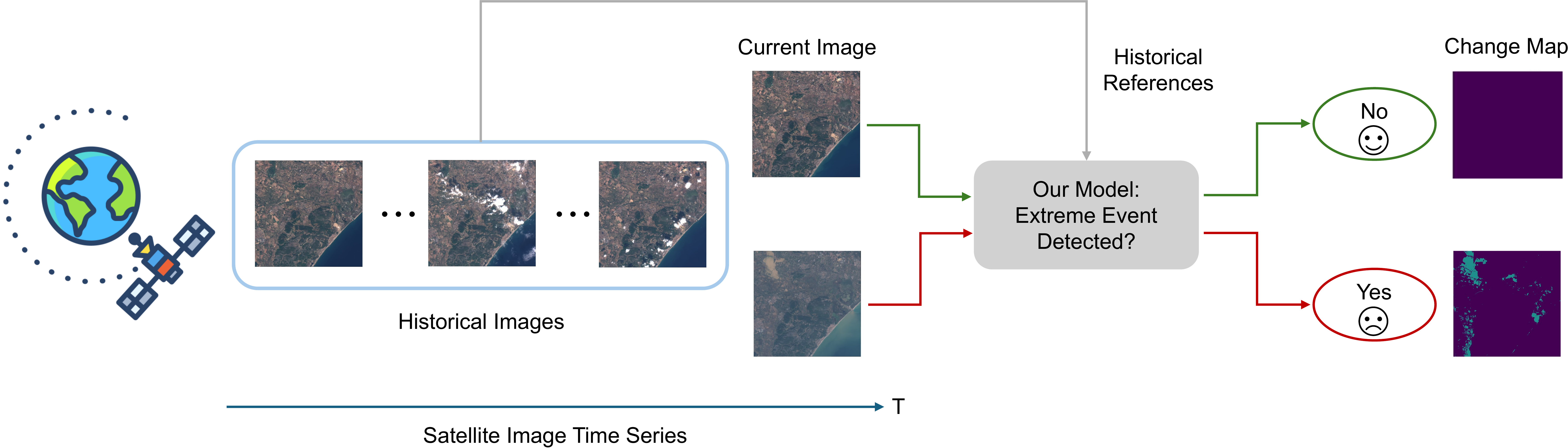} 
\end{center}
\vspace{-5mm}
\caption{
Real-world application of the SITS-Extreme model for extreme event detection. Satellite observations are collected over time, and the latest image is analyzed in comparison with historical data to determine whether an extreme event has occurred. The detection results are then relayed to Earth, enabling rapid response and analysis for disaster management. The figure shows two scenarios: one where no extreme events are detected (using the above current image), and another where extreme events are detected (using the below current image), with the model outputting a change map to highlight affected locations.
}
\label{fig:teaser}
\end{figure*}

Our problem, which focuses on using satellite image time series for accurate extreme event detection, has received limited attention in the literature ~\cite{ruuvzivcka2022ravaen, yadav2024unsupervised}. A closely related area is bi-temporal change detection for damage assessment, which has demonstrated strong performance across diverse local-scale applications ~\cite{zheng2021building,zheng2024towards,gupta2019creating,gerard2024simple}. These methods are theoretically capable of scaling to larger regions, but their application often relies on very high-resolution (VHR) satellite imagery ~\cite{gupta2019creating}. VHR data, typically with resolutions of 0.5 --- 2 m and often provided by commercial satellites, are invaluable for localized analysis but are less readily available for global disaster response due to their limited spatial coverage and accessibility. In contrast, low-to-medium resolution (10 --- 60 m) imagery from open-source satellites like Sentinel-1 and Sentinel-2 offers broader spatial coverage, making it better suited for monitoring large-scale disaster-affected regions.



Another key advantage of open-source satellite imagery, such as Sentinel-1 and Sentinel-2, is their accessibility to offer freely available, globally consistent multi-temporal data. Multi-temporal data provides frequent and systematic observations, enabling the use of multiple pre-disaster images to filter out irrelevant, recurring changes---such as seasonal variations, transient cloud cover, or viewpoint shifts---and to focus on subtle, critical signals that may indicate the onset of extreme events. In contrast, bi-temporal methods, while effective for many applications, may struggle to distinguish between natural, recurring changes, and disaster-relevant changes due to their limited temporal context ~\cite{chen2021remote,chen2022semantic}. By incorporating a longer temporal sequence, multi-temporal approaches excel at differentiating these phenomena, making them particularly well-suited for complex and dynamic scenarios such as disaster detection ~\cite{hafner2024continuous}.


Inspired by these insights, we propose a novel representation learning framework called \textit{SITS-Extreme}, designed to effectively leverage Satellite Image Time Series (SITS) for accurate Extreme events detection. The framework addresses key challenges by integrating contextual information across multiple time steps, enabling it to filter out irrelevant changes and isolate disaster-relevant signals. An illustration of its real-world application is provided in Figure \ref{fig:teaser}. To ensure scalability and accessibility, our approach prioritizes low-to-medium resolution data from open-source satellites, such as Sentinel-1 and Sentinel-2, which offer globally available and freely accessible imagery that is well-suited for large-scale disaster monitoring. However, the framework is resolution-agnostic and can be readily adapted to work with VHR imagery when such data are available.


Our contributions can be summarized as follows:
\begin{itemize}[noitemsep, topsep=0pt]
    \item We address the underexplored challenge of detecting extreme events using satellite image time series, emphasizing its importance and identifying gaps in existing approaches.
    \item We propose the \textit{SITS-Extreme} framework, a novel representation learning approach that leverages multi-temporal satellite data to detect extreme events. The framework integrates contextual information across multiple time steps to filter irrelevant changes and isolate disaster-relevant signals.
    \item We present a comprehensive set of experiments across multiple datasets, including both real-world and synthetic data. Our analysis covers various aspects such as the effectiveness of each component in the architecture, the impact of the number of timesteps, and performance across different disaster types. Our results provide detailed insights and highlight the advantages of multi-temporal approaches over existing baselines.
\end{itemize}

\section{Related Work}
\label{sec:relatedwork}

\begin{figure*}[h]
\begin{center}
\includegraphics[width=1\linewidth]{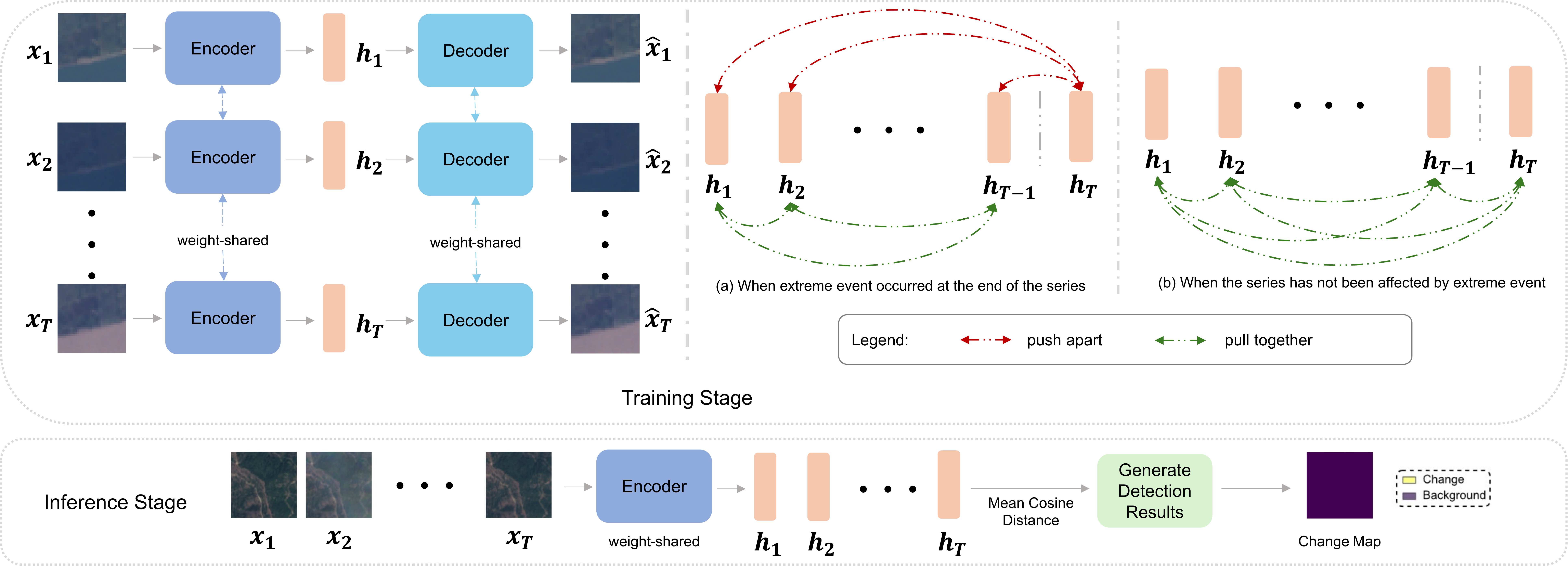} 
\end{center}
\vspace{-5mm}
\caption{Overview of our framework SITS-Extreme. In the training stage, the left part illustrates the autoencoding setup, which learns representations $\{\mathbf{h}_1, \mathbf{h}_2, \ldots, \mathbf{h}_T \}$ from the satellite image time series $\{\mathbf{x}_1, \mathbf{x}_2, \ldots, \mathbf{x}_T \}$, and the right part displays our novel loss functions: (a) the contrastive loss is applied when the time series corresponds to an extreme event, while (b) the consistency loss is used for series unaffected by extreme events. In the inference stage, the learned representations are processed to compute the Mean Cosine Distance (MCD), which determines whether an extreme event has occurred. See Section~\ref{subsec:pipeline} for more details.}
\label{fig:pipeline}
\end{figure*}

Our problem of detecting extreme events using satellite image time series is largely unexplored, with limited prior work.
However, it intersects with several established areas of research, including supervised bi-temporal change detection and spatiotemporal representation learning for satellite imagery. In the following paragraphs, we highlight the relevance of these areas to our approach and discuss the challenges and gaps that prevent existing methods from fully addressing the specific needs of extreme event detection.

\paragraph{Supervised Bi-temporal Change Detection.} Bi-temporal change detection identifies changes between two satellite images and is widely used for disaster damage assessment and environmental monitoring. Recent advances in deep learning, such as fully convolutional networks (FC-EF, FC-Siam-Concat) ~\cite{daudt2018fully, corley2024change} and transformer-based models like BiT and ChangeFormer ~\cite{chen2021remote, bandara2022transformer}, have significantly improved performance, particularly with VHR imagery.

While these methods excel in fine-grained tasks, their reliance on VHR data---often expensive and limited in coverage---makes them less practical for global-scale disaster monitoring. Low-to-medium resolution data, such as imagery from Sentinel-1 and Sentinel-2 satellites, provides open-access coverage with broad geographical and frequent temporal observations, making it well-suited for large-scale applications. However, bi-temporal methods, while effective for detecting changes between two images, are inherently limited in their ability to utilize the richer temporal context provided by multi-temporal satellite data ~\cite{chen2022semantic,hafner2024continuous}. This limitation makes it challenging to filter out recurring natural changes, such as seasonal variations or cloud cover, and to isolate disaster-relevant patterns over time. These challenges highlight the need for approaches that leverage multi-temporal data to address the complexities of extreme event detection more effectively.

\paragraph{Spatio-Temporal Representation Learning for Satellite Imagery.} Self-supervised representation learning for satellite imagery, including remote sensing foundation models, has gained significant attention recently~\cite{lu2024ai,xiao2024foundation,marsocci2024pangaea}. Two prominent approaches in spatiotemporal representation learning are contrastive learning-based and masked autoencoder-based (MAE-based) frameworks. Contrastive learning methods, such as SeCo~\cite{manas2021seasonal} and CaCo~\cite{mall2023change}, leverage temporal differences as augmentations, learning robust spatial representations by contrasting images from different timestamps at the same location. Similarly, MAE-based frameworks like SatMAE~\cite{cong2022satmae} and Scale-MAE~\cite{reed2023scale} extract meaningful representations from multi-spectral, multi-temporal, and multi-scale remote sensing data. For example, SatMAE incorporates temporal embeddings and applies independent masking to image patches across time steps, enabling it to effectively learn general-purpose spatiotemporal representations.

However, these methods are not explicitly tailored to the challenges of extreme events detection. For example, CaCo explicitly focuses on long-term temporal changes, excluding short-term, abrupt changes such as those caused by disasters. As noted in their formulation, disaster-related changes (e.g., landslides or earthquakes) are considered too rare to be useful for their pipeline~\cite{mall2023change}. Similarly, MAE-based frameworks prioritize learning general-purpose representations rather than capturing disaster-specific patterns. 
Extreme events, such as floods and landslides, require identifying subtle but critical changes in highly dynamic environments, often affected by noise such as atmospheric interference, sensor artifacts, or irrelevant changes like seasonal and weather variations. These differ significantly from the recurring or gradual changes typically modeled by these frameworks.
Our approach complements these methods by focusing on extreme events detection, using satellite image time series to isolate disaster-relevant changes while maintaining scalability for large-scale applications.

\section{Method}
\label{sec:method}

We consider the following problem setting for extreme events detection: We have satellite image time series collected over Areas of Interest (AOIs) that have experienced disasters just before the final time step. 
These AOIs span various geographical regions and disaster types, leading to raw satellite images of different sizes. To standardize input and optimize computational resources for satellite deployment, we divide the satellite image time series into fixed-size, non-overlapping spatial patches across all AOIs, creating patch time series for analysis.
Notably, we are interested in designing a mechanism that can benefit from \textit{the availability of multiple pre-disaster images that are unaffected by the event (multi-temporal modeling)} as opposed to prior work that mainly uses the two images before and after the event (bi-temporal modeling). 

\subsection{Problem Formulation}
\label{subsec:formulation}

Suppose we have a patch time series $\mathcal{X} = \{\mathbf{x}_1, \dots, \mathbf{x}_T\}$ of temporal length $T$ and each patch $\mathbf{x}_t \in \mathbb{R}^{C \times P \times P} (t = 1,\dots, T)$, where $C$ refers to the number of channels (bands) in each satellite image and $P$ refers to the patch size. We assume a binary label $y \in \{0, 1\}$ indicating whether the location of $\mathcal{X}$ has been affected by extreme events in the last time step ($t = T$) compared to the previous time steps ($t = 1,..., T-1$) 
Our goal is then to use $\mathcal{X}$ and $y$ to learn patch representations $\mathcal{H} = \{\mathbf{h}_1, \ldots, \mathbf{h}_T \}$ that, given a distance function $\text{D}(\cdot,\cdot)$, can discern new disaster damages\textemdash i.e., $ \text{D}(\mathbf{h}_T, \mathbf{h}_i)\; \forall i < T$, where $\mathbf{h}_T$ corresponds to the representation of the post-disaster image\textemdash from other natural changes, i.e., $ \text{D}(\mathbf{h}_i, \mathbf{h}_j)\; \forall i,j < T$. Specifically, we want that $\text{D}(\mathbf{h}_T, \mathbf{h}_i) \gg \text{D}(\mathbf{h}_i, \mathbf{h}_j)\; \forall i,j < T$ when the patch has been affected by extreme events.

\subsection{Proposed Pipeline: SITS-Extreme}
\label{subsec:pipeline}


\begin{figure*}[ht]
\begin{center}
\includegraphics[width=12.8cm]{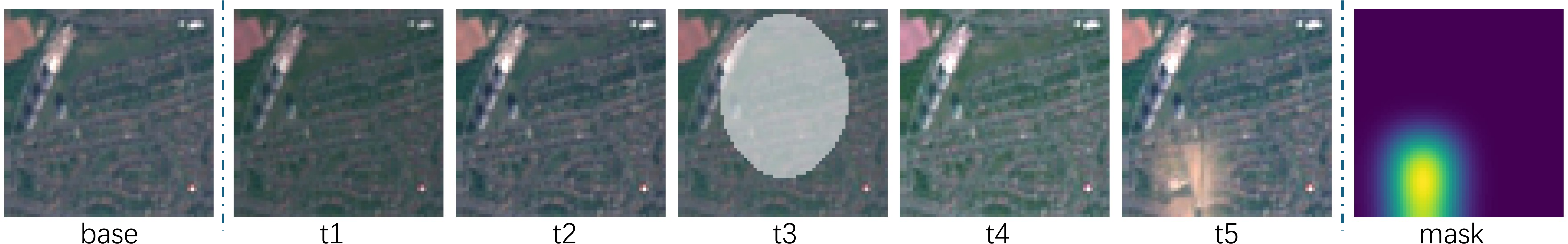} 
\end{center}
\vspace{-5mm}
\caption{An example time series of the generated synthetic dataset. The base image (leftmost) is from the EuroSAT dataset ~\cite{helber2019eurosat}. For $t_5$, we used the CutMix operation with a Gaussian filter to mimic an extreme event. From $t_1$ to $t_5$, we randomly applied the ColorJitter operation and elliptical transparent masks at random positions to mimic seasonal changes and cloud cover. The mask (rightmost) indicates the affected area.}
\label{fig:eurosat}
\end{figure*}

\paragraph{Framework Overview.} To achieve the above goal, we propose a simple yet novel mechanism for accurate extreme events detection, shown in Figure \ref{fig:pipeline}. Specifically, we adopt an autoencoder structure to learn desired latent representations
leveraging the multi-temporal data to discern relevant changes (extreme events) from irrelevant changes (cloud coverage, seasonal changes, etc.).

\paragraph{Patch Time Series Reconstruction.} The overall architecture of our framework follows a transformer-based autoencoder structure \cite{dosovitskiy2020image, carion2020end}, 
forming a mapping between each patch in the time series
$\mathcal{X} \subset \mathbb{R}^{T \times C \times P \times P}$ and its corresponding latent representations $\mathcal{H}$. Two key intuitions motivate the choice of an auto-encoding setup:
\begin{enumerate}[noitemsep, topsep=3pt]
    \item Existing datasets for extreme events detection contain limited data points, often insufficient to learn robust representations. 
    Inspired by Khosla et~al.~\cite{khosla2020supervised}, we employ patch reconstruction as an auxiliary unsupervised task to enhance representation learning and improve generalization~\cite{tschannen2018recent}. 
    \item Foundation models for remote sensing, such as SatMAE~\cite{cong2022satmae} and Scale-MAE~\cite{reed2023scale}, commonly adopt an autoencoder backbone~\cite{lu2024ai}. While not explicitly evaluated for extreme events detection, this design retains compatibility with such models for future integration.
\end{enumerate}

Following notations in Section \ref{subsec:formulation}, 
let $\mathcal{X} = \{\mathbf{x}_1, \dots, \mathbf{x}_T\}$ represent a sequence of patches over time. The reconstruction $\hat{\mathcal{X}} = \{\hat{\mathbf{x}}_1, \dots, \hat{\mathbf{x}}_T\}$ is learned by minimizing the sum of reconstruction losses for all patches in the time series: 
\begin{equation}
    \mathcal{L}_{\text{unified-AE}} = \mathbb{E}_{\mathcal{X}} [\lVert \hat{\mathcal{X}} - \mathcal{X} \rVert ^2_2] + \lambda_{\text{reg}} \mathcal{L}_{\text{reg}}(\mathcal{H})
    \label{eq:unified-ae}
\end{equation}
where $\mathcal{L}_{\text{reg}}(\mathcal{H})$ represents a regularization loss to ensure a well-structured latent space, and $\lambda_{\text{reg}}$ controls its relative weight. 
This formulation allows the model to independently capture patch-level features while leveraging shared weights to learn consistent representations across time steps.
In our approach, the backbone architecture is designed to be flexible, allowing for the use of either a vanilla autoencoder (AE) or a variational autoencoder (VAE).

\paragraph{Task Specific Supervision.} Although the unsupervised training objective helps the architecture learn good latent representations, it does not fully exploit the temporal relationships within the patch time series. As a result, learned latent representations using Equation \ref{eq:unified-ae} alone are insufficient for the task of extreme events detection (See results in Table \ref{tab:abl_loss} for more details). To address this limitation, we introduce task-specific supervision based on the binary label $y$, which indicates whether the location corresponding to this patch time series has been affected by extreme events. This supervision is applied as an additional regularization on the latent spaces. 

When the patch times series has been affected by extreme events (i.e., $y=1$), we aim to achieve two goals: (1) encouraging dissimilar representations between the pre- and post-disaster patches, as indicated by the discrepancy between $\mathbf{h}_T$ and $\{\mathbf{h}_1, \ldots, \mathbf{h}_{T-1} \}$; and (2) ensuring similar representations within the pre-disaster patches, as demonstrated by the consistency among $\{\mathbf{h}_1, \ldots, \mathbf{h}_{T-1} \}$. 
To accomplish these objectives, we design a novel contrastive loss tailored specifically for extreme events detection. Within the affected patch time series $\mathcal{X} = \{\mathbf{x}_1, \dots, \mathbf{x}_T\}$, we consider $\{(\mathbf{x}_i, \mathbf{x}_j) \mid 1 \leq i < j \leq T-1 \}$ as positive pairs and $\{(\mathbf{x}_i, \mathbf{x}_T) \mid 1 \leq i < T\}$ as negative pairs, resulting in $\frac{{(T-1)(T-2)}}{2}$ positive pairs and $T-1$ negative pairs. 
Unlike the standard NT-Xent contrastive loss in SimCLR~\cite{chen2020simple}, which emphasizes balancing a single positive pair with many negative pairs, our method instead uses more positive pairs than negative ones. 
This aligns with the nature of extreme events detection, where both capturing consistent patterns across positive pairs (e.g., seasonal variations) and maximizing separation between negative pairs (e.g., pre- vs. post-disaster) are crucial for improving sensitivity to disaster-relevant signals.
By pushing apart each negative pair, we ensure their distances are significantly greater than those of all positive pairs, distinguishing disaster-induced changes from recurring natural variations.
This gives us the contrastive loss for an affected patch time series:

\begin{equation}
\begin{split}
    l_\text{contra} = 
    \frac{1}{T-1} (-\sum_{a=1}^{T-1} \text{log} \frac{\text{exp}(\text{D}(\mathbf{h}_a, \mathbf{h}_T))}{\sum_{b=1}^{T-1} \mathds{1}_{[b \neq a]} \text{exp}(\text{D}(\mathbf{h}_a, \mathbf{h}_b))})
\end{split}
\end{equation}

When the patch time series has not been affected by extreme events (i.e., $y=0$), we aim to maintain similar representations throughout the entire patch time series, as indicated by the consistency among $\{\mathbf{h}_1, \ldots, \mathbf{h}_T \}$. Given that we have all possible $\frac{T(T-1)}{2}$ pairs $\{(\mathbf{x}_i, \mathbf{x}_j) \mid 1 \leq i < j \leq T \}$, this gives us the consistency loss for a non-affected patch time series:

\begin{equation}
    l_\text{consist} = 
    \frac{1}{T(T-1)}\sum_{a=1}^T \sum_{b=1}^T \mathds{1}_{[b \neq a]} \text{D}(\mathbf{h}_a, \mathbf{h}_b)
\end{equation}

\paragraph{Summary.} Given $N$ training samples of patch time series, we reach the training objective function of our framework:
\begin{equation}
\begin{split}
    \mathcal{L} = & \mathcal{L}_{\text{unified-AE}} + \lambda \frac{\mathds{1}_{\{y=1\}}} {\sum_{s=1}^N {\mathds{1}_{\{y=1\}}} } l_\text{contra} (s) \\
    & + \mu \frac{\mathds{1}_{\{y=0\}}} {\sum_{s=1}^N {\mathds{1}_{\{y=0\}}} } l_\text{consist} (s)
\end{split}
\end{equation}
where $\lambda$ and $\mu$ denote the loss balancing weights.

For the distance function $\text{D}(.,.)$, we employ cosine distance in our embedding space to maximize dissimilarity:
\begin{equation}
    \text{D}(\mathbf{p}, \mathbf{q}) = 1 - \lvert \frac{\mathbf{p} \cdot \mathbf{q}}{\lVert \mathbf{p} \rVert \lVert \mathbf{q} \rVert} \rvert
\label{eq:cd}
\end{equation}
This choice is motivated by the fact that cosine distance is both upper- and lower-bounded, making it particularly effective for comparing latent representations by focusing on their direction rather than their magnitude.

\paragraph{Inference Phase. } During inference, given a satellite patch time series $\mathcal{X} = \{\mathbf{x}_1, \ldots, \mathbf{x}_T\}$, our trained model processes the input patches using the weight-shared vision transformer encoder to produce the learned representations $\mathcal{H} = \{\mathbf{h}_1, \ldots, \mathbf{h}_T\}$. These representations capture spatiotemporal features for each patch in the time series.

To detect whether the patch time series has been affected by an extreme event, we compute the \textit{Mean Cosine Distance (MCD)} between the pre-disaster representations $\{\mathbf{h}_1, \ldots, \mathbf{h}_{T-1}\}$ and the post-disaster representation $\mathbf{h}_T$: 
\begin{equation}
    \text{MCD} = \frac{1}{T-1} \sum_{i=1}^{T-1} D(\mathbf{h}_i, \mathbf{h}_T)
    \label{eq:mcd} 
\end{equation}

The MCD quantifies the degree of change, serving as an indicator of whether an extreme event has occurred. A high MCD value indicates significant differences, suggesting disaster impact, while lower values imply minimal or no event-related changes. To produce the final binary detection result, the MCD is compared against a predefined threshold $\tau$, determined during validation: 
\begin{equation}
    \hat{y} = 
    \begin{cases}
        1, & \text{if } \text{MCD} \geq \tau, \\
        0, & \text{otherwise.}         
    \end{cases}
\end{equation}

\section{Experiments}
\label{sec:experiment}
In this section, we conduct experiments on both synthetic and real-world datasets. The RaV{\AE}n dataset~\cite{ruuvzivcka2022ravaen} is the only available open-source real-world satellite image time series dataset for extreme events detection. 
To complement this, we first construct a synthetic dataset designed to validate the core assumption of our method: that disaster-relevant signals can be effectively discerned from irrelevant, recurring changes in satellite image time series. This synthetic dataset simulates multi-temporal dynamics in a more balanced and less complex setting than real-world scenarios, enabling us to evaluate whether our method captures relevant patterns across time steps in a controlled setting. We then apply our method to the real-world RaV{\AE}n dataset to demonstrate its performance in practical scenarios.


\begin{figure*}[ht]
\begin{center}
\includegraphics[width=12.8cm]{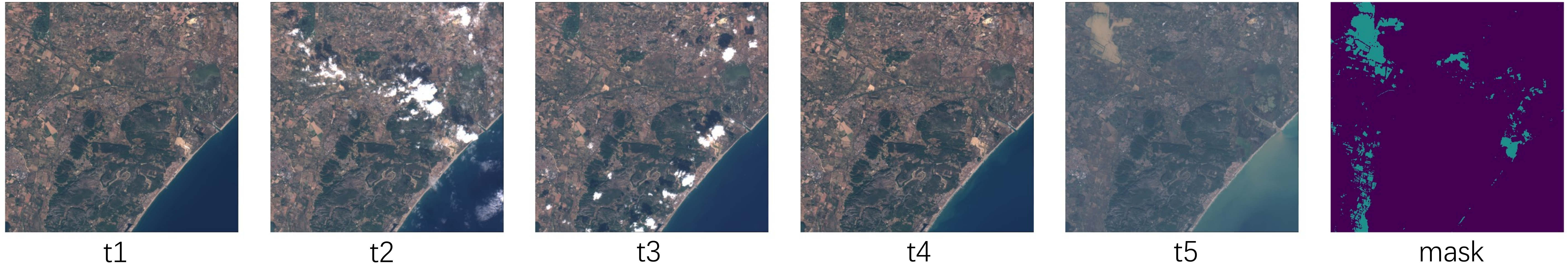} 
\end{center}
\vspace{-5mm}
\caption{A scene from the real-world RaV{\AE}n dataset illustrates a satellite patch time series affected by flood. $t_1$ through $t_4$ are pre-disaster images, and $t_5$ is the post-disaster image, all captured by the Sentinel-2 satellite. The mask (rightmost) highlights the flooded area.}
\label{fig:ravaen}
\end{figure*}

\subsection{Datasets}
\paragraph{Synthetic Dataset. } 
One defining characteristic of a satellite image time series is that images captured at different timesteps originate from the same geolocation, ensuring shared geospatial information across the series. Leveraging this characteristic, we constructed a synthetic dataset by selecting an existing satellite dataset as the ``base image" for the time series. We then applied a combination of task-irrelevant and task-relevant transformations to simulate both natural changes and extreme events. 
For the ``base image", we selected the EuroSAT dataset, which contains 27,000 individual Sentinel-2 satellite images spanning 10 land use and land cover (LULC) classes~\cite{helber2019eurosat}. Unlike pre-existing time series datasets, EuroSAT consists of standalone images, ensuring that our synthetic time series are free from uncontrolled changes or unforeseen extreme events present in real-world sequences.
To simulate task-irrelevant changes, we applied color jittering to mimic seasonal variations, and we added a random elliptical transparent mask to mimic cloud cover. To simulate task-relevant changes - extreme events, we applied the CutMix operation with a Gaussian filter to the base image, in which a region of the base image is replaced with a patch from another image of a different LULC class. The process of generating the synthetic datasets, detailed in Algorithm \ref{alg:synthetic} in Section \ref{appendix:alg} of the Appendix, resulted in 27,000 satellite image time series, each consisting of 5 timesteps. We divided this dataset into training, validation, and testing sets in a 7:1:2 ratio. An example of the generated dataset is shown in Figure \ref{fig:eurosat}.

\paragraph{Real-world Datasets.} The real-world dataset RaV{\AE}n comprises 19 AOIs captured by the Sentinel-2 satellite, covering four disaster classes: fire burn scars, floods, hurricanes, and landslides \cite{ruuvzivcka2022ravaen}. These events were carefully identified through an extensive review of Sentinel-2 records, supported by the Copernicus Emergency Management Service (CEMS). 
Each event in the dataset consists of a satellite image time series with 5 timesteps, where the first four images are taken prior to the disaster, and the fifth image reflects changes after the event.
An example AOI from the dataset is shown in Figure \ref{fig:ravaen}, demonstrating the complexity of the task. Given the variations in AOIs and disaster types, the raw satellite images often come in different sizes. To standardize the data, we segmented the RaV{\AE}n dataset into non-overlapping 64x64 patches, which allowed for a patch-level focus in our extreme events detection task. Binary labels (0 or 1) were assigned to each satellite patch time series based on the original change mask, indicating whether the patch had been affected by an extreme event. The preprocessed dataset was then divided into training, validation, and testing sets in a 7:1:2 ratio. Upon further exploration, we observed significant class imbalance across the different disaster types, as shown in Table \ref{tab:ravaen_statistics}.


\begin{table}[h]
\footnotesize
\begin{center}
\begin{tabular}{@{}cccc@{}}
\toprule
 & \# of AOIs & \# of patches & Positive ratio \\ \midrule
Fires & 5 & 8303 & 58.32\%  \\
Floods & 4 & 3069 & 32.13\%  \\
Hurricanes & 5 & 3842 & 36.23\%  \\
Landslides & 5 & 221 & 29.41\% \\ \bottomrule
\end{tabular}
\caption{
The RaV{\AE}n dataset statistics: 
\textit{Positive ratio} refers to the proportion of patches with the class label of 1 (patches have been affected by extreme events) relative to the total number of patches for each disaster type.
}
\label{tab:ravaen_statistics}
\end{center}
\end{table}

\vspace{-5mm}

\begin{table*}[t!]
\footnotesize
\begin{center}
\begin{tabular}{@{}cccccc@{}}
\toprule
Dataset & Method & AP $\uparrow$ & F1 $\uparrow$ & Precision $\uparrow$ & Recall $\uparrow$ \\ \midrule
\multirow{6}{*}{Synthetic} 
& 5-Step Bi-SiamConcat &           \underline{0.9608 $\pm$  0.0065} &           \underline{0.8804 $\pm$  0.0088} &           \underline{0.9017 $\pm$  0.0195} &           0.8602 $\pm$  0.0050 \\ 
& 5-Step Bi-SiamDiff &           0.8800 $\pm$  0.0022 &           0.8047 $\pm$  0.0005 &           0.7208 $\pm$  0.0103 &           \underline{0.9112 $\pm$  0.0155} \\ 
& 5-Step Multi-SiamConcat &           0.9519 $\pm$  0.0005 &           0.8640 $\pm$  0.0050 &           0.8598 $\pm$  0.0471 &           0.8743 $\pm$  0.0545 \\ 
& 5-Step Multi-SiamDiff &           0.8220 $\pm$  0.0011 &           0.7837 $\pm$  0.0001 &           0.6645 $\pm$  0.0007 &           \textbf{0.9550 $\pm$  0.0011} \\ 
& \textbf{SITS-Extreme-AE} &           \textbf{0.9646 $\pm$  0.0087} &           \textbf{0.8999 $\pm$  0.0173} &           \textbf{0.9276 $\pm$  0.0073} &           0.8741 $\pm$  0.0277 \\ 
& \textbf{SITS-Extreme-VAE} &           0.9467 $\pm$  0.0018 &           0.8632 $\pm$  0.0017 &           0.8707 $\pm$  0.0102 &           0.8559 $\pm$  0.0071 \\ \midrule
\multirow{6}{*}{ RaV{\AE}n}
& 5-Step Bi-SiamConcat &           0.7993 $\pm$  0.0038 &           0.7801 $\pm$  0.0014 &           0.7205 $\pm$  0.0110 &           \underline{0.8508 $\pm$  0.0151} \\ 
& 5-Step Bi-SiamDiff &           0.7917 $\pm$  0.0106 &           0.7504 $\pm$  0.0057 &           0.6942 $\pm$  0.0126 &           0.8172 $\pm$  0.0197 \\ 
& 5-Step Multi-SiamConcat &           0.8003 $\pm$  0.0097 &           0.7759 $\pm$  0.0032 &           0.7205 $\pm$  0.0257 &           0.8429 $\pm$  0.0310 \\ 
& 5-Step Multi-SiamDiff &           0.7481 $\pm$  0.0069 &           0.7312 $\pm$  0.0048 &           0.6354 $\pm$  0.0164 &           \textbf{0.8619 $\pm$  0.0170} \\ 
& \textbf{SITS-Extreme-AE} &           \underline{0.9014 $\pm$  0.0049} &           \underline{0.8153 $\pm$  0.0140} &          \textbf{0.8822 $\pm$  0.0152} &           0.7579 $\pm$  0.0143 \\
& \textbf{SITS-Extreme-VAE} &           \textbf{0.9069 $\pm$  0.0021} &           \textbf{0.8473 $\pm$  0.0021} &           \underline{0.8622 $\pm$  0.0166} &           0.8336 $\pm$  0.0175 \\ \midrule
RaV{\AE}n &  2-Step Multi-SiamConcat &           0.7195 $\pm$ 0.0051 &           0.7479 $\pm$ 0.0009 &           0.6490 $\pm$ 0.0105 &           0.8831 $\pm$ 0.0204 \\
\bottomrule
\end{tabular}
\caption{\textbf{Main comparison results on the synthetic dataset and the real-world RaV{\AE}n dataset.} We \textit{reimplement} 2-Step Multi-SiamConcat, 5-Step Bi-SiamConcat, 5-Step Bi-SiamDiff, 5-Step Multi-SiamConcat, 5-Step Multi-SiamDiff into the \textit{same learning setup} and the same transformer-based architecture. We used \textit{same hyperparameter optimization budget and mechanism} for all the baseline methods and ours. 
Mean test accuracy and standard deviation are reported from three runs. Statistically significant top performers are boldfaced, and the second-best performers are underlined.}
\label{tab:main_results}
\vspace{-6mm}
\end{center}
\end{table*}

\subsection{Baseline Models}
Given the novelty of our task, there are no established benchmark models for direct comparison. To address this, we identified and adapted suitable baselines from the field of bi-temporal change detection. For a fair comparison, we prioritized models with strong performance in this area, particularly those that, like our method, emphasize feature interaction mechanisms. Previous works ~\cite{corley2024change} and ~\cite{fang2023changer} have demonstrated that feature interaction is crucial for improving performance in change detection tasks. 
By focusing on this aspect rather than backbone architecture, we ensure that any observed performance differences reflect the models’ ability to handle the task.

Following the conclusions of ~\cite{corley2024change}, SiamConcat and SiamDiff ~\cite{daudt2018fully} have shown strong performance on the bi-temporal LEVIR-CD and WHU-CD datasets, benefiting from their simple yet effective feature interaction mechanisms. 
We adapted these methods, originally designed for bi-temporal inputs, to handle satellite patch time series $\mathcal{X} = \{\mathbf{x}_1, \dots, \mathbf{x}_T\}$ in the following ways:


\begin{itemize}[noitemsep, topsep=0pt] 
\item \textbf{Bi-temporal Strategy}: To adapt bi-temporal methods for our task, the model was extended to construct $T-1$ bi-temporal pairs by pairing each $\mathbf{x}_i (i = 1, \dots, T-1)$ with $\mathbf{x}_T$. The original SiamConcat and SiamDiff methods were applied to each pair, and the outputs were aggregated to produce the final prediction. This adaptation resulted in \textit{5-Step Bi-SiamConcat} and \textit{5-Step Bi-SiamDiff} baselines, explicitly leveraging all five timesteps.
\item \textbf{Multi-temporal Strategy}: Here, the entire satellite patch time series $\mathcal{X}$ was treated as a whole. For Multi-SiamConcat, we extended the concatenation mechanism to operate across all timesteps by concatenating feature maps $\{\mathbf{h}_1, \dots, \mathbf{h}_T\}$. For Multi-SiamDiff, we adapted the differencing mechanism to compute pairwise differences between $\mathbf{h}_T$ and all previous timesteps $\{\mathbf{h}_1, \dots, \mathbf{h}_{T-1}\}$, aggregating these differences to generate the final prediction. This resulted in the \textit{5-Step Multi-SiamConcat} and \textit{5-Step Multi-SiamDiff} explicitly utilizing all five timesteps.
\end{itemize}

To ensure fair comparisons, all baseline models and our method were implemented using the same backbone architectures. Details of their implementations are provided in Section \ref{appendix:imp} of the Appendix.
\subsection{Evaluation Metrics}



The resulting MCD is used to calculate the following evaluation metrics, where AP is the primary metric, providing a threshold-independent measure of performance. F1, precision, and recall are determined by selecting the optimal threshold via grid search on the testing set to show the best possible performance.

\begin{itemize}[noitemsep, topsep=0pt]
\item \textbf{Average Precision (AP)}: The area under the precision-recall curve, offers a threshold-independent measure of performance.

\item \textbf{F1 Score}: A balance between precision and recall, particularly useful for imbalanced datasets where minimizing both false alarms and missed events is crucial.

\item \textbf{Precision}: The proportion of correctly identified extreme events (higher precision means fewer false alarms).

\item\textbf{Recall}: The model's ability to detect actual extreme events (higher recall meaning fewer missed events).

\end{itemize}


\subsection{Main Comparative Study}

\begin{table*}[t!]
\footnotesize
\begin{center}
\begin{tabular}{@{}ccccccc@{}}
\toprule
$\mathcal{L}_{\text{unified-AE}}$ & $l_\text{contra}$ & $l_\text{consist}$ & AP $\uparrow$ & F1 $\uparrow$ & Precision $\uparrow$ & Recall $\uparrow$ \\ \midrule
\ding{55} & \ding{51} & \ding{51} & / & / & / & / \\
\ding{51} & \ding{55} & \ding{55} &           0.5006 $\pm$ 0.0050 &           0.6461 $\pm$ 0.0012 &           0.4909 $\pm$ 0.0056 &           0.9455 $\pm$ 0.0180 \\
\ding{51} & \ding{51} & \ding{55} &           0.6679 $\pm$ 0.0045 &           0.6876 $\pm$ 0.0029 &           0.5768 $\pm$ 0.0011 &           0.8510 $\pm$ 0.0113 \\
\ding{51} & \ding{51} & \ding{51} &           0.9069 $\pm$ 0.0021 &           0.8473 $\pm$ 0.0021 &           0.8622 $\pm$ 0.0166 &           0.8336 $\pm$ 0.0175 \\
\bottomrule
\end{tabular}
\caption{\textbf{Ablation study on the effectiveness of each component in the SITS-Extreme framework on the real-world RaV{\AE}n dataset.} This study evaluates the importance of three key components: $\mathcal{L}_{\text{unified-AE}}$, $l_\text{contra}$, and $l_\text{consist}$. }
\label{tab:abl_loss}
\vspace{-5mm}
\end{center}
\end{table*}

\begin{table*}[t!]
\footnotesize
\begin{center}
\begin{tabular}{@{}ccccc@{}}
\toprule
Disaster & AP $\uparrow$ & F1 $\uparrow$ & Precision $\uparrow$ & Recall $\uparrow$ \\ \midrule
fire &             0.9308 $\pm$  0.0033 &             0.8858 $\pm$  0.0021 &             0.9220 $\pm$  0.0058 &             0.8525 $\pm$  0.0068 \\
flood &             0.7353 $\pm$  0.0103 &             0.6945 $\pm$  0.0195 &             0.6089 $\pm$  0.0429 &             0.8135 $\pm$  0.0314 \\
hurricane &             0.9491 $\pm$  0.0051 &             0.8709 $\pm$  0.0308 &             0.9599 $\pm$  0.0039 &             0.7983 $\pm$  0.0500 \\
landslide &             0.4395 $\pm$  0.0427 &             0.3627 $\pm$  0.0730 &             0.3288 $\pm$  0.0400 &             0.4242 $\pm$  0.1545 \\
\bottomrule
\end{tabular}
\caption{\textbf{Ablation study on disaster types in the real-world RaV{\AE}n dataset.} Due to the imbalanced distribution of disaster types in the RaV{\AE}n dataset, we evaluate the performance of our best-performing model (SITS-Extreme-VAE) on four specific disaster types: fire, flood, hurricane, and landslide. The results highlight variations in model performance across different disaster categories.}
\label{tab:abl_disaster}
\vspace{-5mm}
\end{center}
\end{table*}

We implement our framework with both vanilla AE and VAE backbones, thus we call them SITS-Extreme-AE and SITS-Extreme-VAE. Table \ref{tab:main_results} presents the main comparison results between our proposed models and the baseline models on both the synthetic and real-world RaV{\AE}n datasets. To ensure a fair comparison, all methods were re-implemented with the same learning setup, architecture, and hyperparameter optimization process.

\paragraph{Results on Synthetic Dataset.} 
Our model, SITS-Extreme-AE, achieves the highest performance across most metrics, with an AP of 0.9646 and an F1 score of 0.8999, demonstrating a strong balance between precision and recall. 
Compared to the best-performing baseline, 5-Step Bi-SiamConcat, our method achieves a 0.38\% improvement in AP and a 1.95\% improvement in F1 score.
Among the baseline models, methods utilizing the concatenation operation (e.g., 5-Step Bi-SiamConcat) tend to perform better for multi-temporal inputs. Although Multi-SiamDiff achieves the highest recall (0.9550), it suffers from significantly lower precision (0.6645), indicating a higher rate of false alarms, negatively affecting its overall performance.

\paragraph{Results on Real-world RaV{\AE}n Dataset.} 
On the real-world RaV{\AE}n dataset, our model, SITS-Extreme-VAE, outperforms all other methods with an AP of 0.9069 and an F1 score of 0.8473. The precision of 0.8622 underscores its ability to minimize false alarms while maintaining strong recall. Compared to the best baseline, 5-Step Multi-SiamConcat, 
our method shows a 10.66\% improvement in AP and a 7.14\% improvement in F1 score. Given the increased difficulty of the real-world dataset compared to the synthetic dataset, these results demonstrate the superiority of our method in handling satellite image time series. Although 5-Step Multi-SiamDiff achieves the highest recall (0.8619), its much lower precision (0.6354) suggests it is prone to much more false positives.


\paragraph{Effectiveness of Satellite Patch Time Series.}
To evaluate the impact of incorporating more timesteps, we selected the best-performing baseline, 5-Step Multi-SiamConcat, on the RaV{\AE}n dataset for further analysis. We modified its architecture to use only the last two timesteps, $\mathbf{x}_{T-1}$ and $\mathbf{x}_T$, replicating the original bi-temporal setup. This variant, referred to as 2-Step Multi-SiamConcat, provides a simpler comparison point. The results clearly demonstrate the advantages of using more timesteps,
with an 8.08\% improvement in AP and a 2.8\% improvement in F1 score for 5-Step Multi-SiamConcat compared to its 2-step counterpart.
The higher precision of the 5-step configuration (0.7205 vs. 0.6490) also underscores its ability to reduce false positives. These results highlight the importance of leveraging additional pre-disaster timesteps to provide valuable contextual information for distinguishing disaster-relevant signals from recurring natural changes, improving overall performance in extreme events detection.

\subsection{Ablation Study}
To validate the design choices of our framework, we conducted two ablation studies on the real-world RaV{\AE}n dataset. 
First, we investigate the contribution of each component in our framework. Second, we evaluate the effectiveness of our method across different disaster types, given the multi-disaster nature of the dataset.
\vspace{-2mm}

\paragraph{How Effective Is Each Loss Term?}
As shown in Table \ref{tab:abl_loss}, the first row represents the scenario where the autoencoding setup is removed. Without this setup, the model becomes unstable during training and fails to converge, resulting in no meaningful performance metrics. Subsequently, we observe the significant performance degradation when the model is trained without the task-specific supervision  $l_\text{consist}$ and $l_\text{contra}$. The model's performance drops considerably when the autoencoding setup is used alone without these two losses. Notably, adding $l_\text{consist}$ leads to a greater improvement compared to $l_\text{contra}$, indicating that leveraging consistency information across multiple pre-disaster images helps filter out task-irrelevant changes more effectively. Thus, every component of our framework contributes to enhancing the overall performance of the proposed method.


\vspace{-2mm}

\paragraph{How does our method perform in different disaster types?} 

In this ablation study, we examine the performance of SITS-Extreme-VAE on different types in the RaV{\AE}n dataset, focusing on fire, flood, hurricane, and landslide, as shown in Table \ref{tab:abl_disaster}. The model achieves its best performance on fire (0.9308 AP) and hurricane (0.9491 AP), reflecting its ability to handle these better-represented disaster types. In contrast, the model's performance drops on floods (0.7353 AP) and significantly on landslides (0.4395 AP), where the number of patches and positive ratio are lower (detailed in Table \ref{tab:ravaen_statistics}). The lower precision in floods (0.6089) and landslides (0.3288) suggests the model struggles with false positives, likely due to the limited number of samples in this category. 
This underscores the impact of data imbalance on performance and suggests the need for refinement for underrepresented events like landslides.

\section{Conclusion and Future Work}
\label{sec:conclusion}

In this work, we proposed SITS-Extreme, a novel framework that effectively leverages satellite image time series for the accurate detection of extreme events. By incorporating multiple pre-disaster observations, our approach addresses the limitations of traditional bi-temporal methods, which often struggle to filter out irrelevant changes and capture subtle event signals. Through experiments on both real-world and synthetic datasets, we demonstrated the effectiveness of SITS-Extreme in providing a robust solution for disaster response. For future work, we aim to expand our small-scale evaluation to large-scale datasets that better reflect real-world disaster response needs. Additionally, we plan to extend SITS-Extreme to handle multi-modal data, such as combining optical and SAR imagery, to address challenges like cloud cover and data availability. Using pretrained multi-modal encoders, such as the Dynamic One-For-All (DOFA) model, could enable integration of pre- and post-event data from different satellite sources ~\cite{xiong2024neural}.



\paragraph{Acknowledgements.}
This work is funded by Digital Futures in the project EO-AI4GlobalChange. All experiments were performed using the supercomputing resource Berzelius provided by the National Supercomputer Centre at Linkoping University and the Knut and Alice Wallenberg Foundation. Heng Fang thanks Erik Englesson, Adam Stewart, Dino Ienco, Zhuo Zheng, Sebastian Gerard, Ling Li, and Sebastian Hafner for their feedback on improving the presentation of this paper.

\clearpage
{\small
\bibliographystyle{ieee_fullname}
\bibliography{egbib}

\begin{thebibliography}{10}\itemsep=-1pt

\bibitem{bandara2022transformer}
Wele Gedara~Chaminda Bandara and Vishal~M Patel.
\newblock A transformer-based siamese network for change detection.
\newblock In {\em IGARSS 2022-2022 IEEE International Geoscience and Remote Sensing Symposium}, pages 207--210. IEEE, 2022.

\bibitem{carion2020end}
Nicolas Carion, Francisco Massa, Gabriel Synnaeve, Nicolas Usunier, Alexander Kirillov, and Sergey Zagoruyko.
\newblock End-to-end object detection with transformers.
\newblock In {\em European conference on computer vision}, pages 213--229. Springer, 2020.

\bibitem{chen2021remote}
Hao Chen, Zipeng Qi, and Zhenwei Shi.
\newblock Remote sensing image change detection with transformers.
\newblock {\em IEEE Transactions on Geoscience and Remote Sensing}, 60:1--14, 2021.

\bibitem{chen2022semantic}
Hao Chen, Yifan Zao, Liqin Liu, Song Chen, and Zhenwei Shi.
\newblock Semantic decoupled representation learning for remote sensing image change detection.
\newblock In {\em IGARSS 2022-2022 IEEE International Geoscience and Remote Sensing Symposium}, pages 1051--1054. IEEE, 2022.

\bibitem{chen2020simple}
Ting Chen, Simon Kornblith, Mohammad Norouzi, and Geoffrey Hinton.
\newblock A simple framework for contrastive learning of visual representations.
\newblock In {\em International conference on machine learning}, pages 1597--1607. PMLR, 2020.

\bibitem{cong2022satmae}
Yezhen Cong, Samar Khanna, Chenlin Meng, Patrick Liu, Erik Rozi, Yutong He, Marshall Burke, David Lobell, and Stefano Ermon.
\newblock Satmae: Pre-training transformers for temporal and multi-spectral satellite imagery.
\newblock {\em Advances in Neural Information Processing Systems}, 35:197--211, 2022.

\bibitem{corley2024change}
Isaac Corley, Caleb Robinson, and Anthony Ortiz.
\newblock A change detection reality check.
\newblock {\em arXiv preprint arXiv:2402.06994}, 2024.

\bibitem{cred2023}
Centre for Research on the Epidemiology of~Disasters (CRED) and US~Agency Int.~Dev. (USAID).
\newblock Disaster year in review 2023., 2024.

\bibitem{daudt2018fully}
Rodrigo~Caye Daudt, Bertr Le~Saux, and Alexandre Boulch.
\newblock Fully convolutional siamese networks for change detection.
\newblock In {\em 2018 25th IEEE international conference on image processing (ICIP)}, pages 4063--4067. IEEE, 2018.

\bibitem{dosovitskiy2020image}
Alexey Dosovitskiy, Lucas Beyer, Alexander Kolesnikov, Dirk Weissenborn, Xiaohua Zhai, Thomas Unterthiner, Mostafa Dehghani, Matthias Minderer, Georg Heigold, Sylvain Gelly, et~al.
\newblock An image is worth 16x16 words: Transformers for image recognition at scale.
\newblock {\em arXiv preprint arXiv:2010.11929}, 2020.

\bibitem{fang2023changer}
Sheng Fang, Kaiyu Li, and Zhe Li.
\newblock Changer: Feature interaction is what you need for change detection.
\newblock {\em IEEE Transactions on Geoscience and Remote Sensing}, 61:1--11, 2023.

\bibitem{gerard2024simple}
Sebastian Gerard, Paul Borne-Pons, and Josephine Sullivan.
\newblock A simple, strong baseline for building damage detection on the xbd dataset.
\newblock {\em arXiv preprint arXiv:2401.17271}, 2024.

\bibitem{gupta2019creating}
Ritwik Gupta, Bryce Goodman, Nirav Patel, Ricky Hosfelt, Sandra Sajeev, Eric Heim, Jigar Doshi, Keane Lucas, Howie Choset, and Matthew Gaston.
\newblock Creating xbd: A dataset for assessing building damage from satellite imagery.
\newblock In {\em Proceedings of the IEEE/CVF conference on computer vision and pattern recognition workshops}, pages 10--17, 2019.

\bibitem{hafner2024continuous}
Sebastian Hafner, Heng Fang, Hossein Azizpour, and Yifang Ban.
\newblock Continuous urban change detection from satellite image time series with temporal feature refinement and multi-task integration.
\newblock {\em arXiv preprint arXiv:2406.17458}, 2024.

\bibitem{he2022masked}
Kaiming He, Xinlei Chen, Saining Xie, Yanghao Li, Piotr Doll{\'a}r, and Ross Girshick.
\newblock Masked autoencoders are scalable vision learners.
\newblock In {\em Proceedings of the IEEE/CVF conference on computer vision and pattern recognition}, pages 16000--16009, 2022.

\bibitem{helber2019eurosat}
Patrick Helber, Benjamin Bischke, Andreas Dengel, and Damian Borth.
\newblock Eurosat: A novel dataset and deep learning benchmark for land use and land cover classification.
\newblock {\em IEEE Journal of Selected Topics in Applied Earth Observations and Remote Sensing}, 12(7):2217--2226, 2019.

\bibitem{khosla2020supervised}
Prannay Khosla, Piotr Teterwak, Chen Wang, Aaron Sarna, Yonglong Tian, Phillip Isola, Aaron Maschinot, Ce Liu, and Dilip Krishnan.
\newblock Supervised contrastive learning.
\newblock {\em Advances in neural information processing systems}, 33:18661--18673, 2020.

\bibitem{lu2024ai}
Siqi Lu, Junlin Guo, James~R Zimmer-Dauphinee, Jordan~M Nieusma, Xiao Wang, Parker VanValkenburgh, Steven~A Wernke, and Yuankai Huo.
\newblock Ai foundation models in remote sensing: A survey.
\newblock {\em arXiv preprint arXiv:2408.03464}, 2024.

\bibitem{mall2023change}
Utkarsh Mall, Bharath Hariharan, and Kavita Bala.
\newblock Change-aware sampling and contrastive learning for satellite images.
\newblock In {\em Proceedings of the IEEE/CVF Conference on Computer Vision and Pattern Recognition}, pages 5261--5270, 2023.

\bibitem{manas2021seasonal}
Oscar Manas, Alexandre Lacoste, Xavier Gir{\'o}-i Nieto, David Vazquez, and Pau Rodriguez.
\newblock Seasonal contrast: Unsupervised pre-training from uncurated remote sensing data.
\newblock In {\em Proceedings of the IEEE/CVF International Conference on Computer Vision}, pages 9414--9423, 2021.

\bibitem{marsocci2024pangaea}
Valerio Marsocci, Yuru Jia, Georges~Le Bellier, David Kerekes, Liang Zeng, Sebastian Hafner, Sebastian Gerard, Eric Brune, Ritu Yadav, Ali Shibli, et~al.
\newblock Pangaea: A global and inclusive benchmark for geospatial foundation models.
\newblock {\em arXiv preprint arXiv:2412.04204}, 2024.

\bibitem{paszke2019pytorch}
Adam Paszke, Sam Gross, Francisco Massa, Adam Lerer, James Bradbury, Gregory Chanan, Trevor Killeen, Zeming Lin, Natalia Gimelshein, Luca Antiga, et~al.
\newblock Pytorch: An imperative style, high-performance deep learning library.
\newblock {\em Advances in neural information processing systems}, 32, 2019.

\bibitem{reed2023scale}
Colorado~J Reed, Ritwik Gupta, Shufan Li, Sarah Brockman, Christopher Funk, Brian Clipp, Kurt Keutzer, Salvatore Candido, Matt Uyttendaele, and Trevor Darrell.
\newblock Scale-mae: A scale-aware masked autoencoder for multiscale geospatial representation learning.
\newblock In {\em Proceedings of the IEEE/CVF International Conference on Computer Vision}, pages 4088--4099, 2023.

\bibitem{ruuvzivcka2022ravaen}
V{\'\i}t R{\u{u}}{\v{z}}i{\v{c}}ka, Anna Vaughan, Daniele De~Martini, James Fulton, Valentina Salvatelli, Chris Bridges, Gonzalo Mateo-Garcia, and Valentina Zantedeschi.
\newblock Rav{\ae}n: unsupervised change detection of extreme events using ml on-board satellites.
\newblock {\em Scientific reports}, 12(1):16939, 2022.

\bibitem{tschannen2018recent}
Michael Tschannen, Olivier Bachem, and Mario Lucic.
\newblock Recent advances in autoencoder-based representation learning.
\newblock {\em arXiv preprint arXiv:1812.05069}, 2018.

\bibitem{xiao2024foundation}
Aoran Xiao, Weihao Xuan, Junjue Wang, Jiaxing Huang, Dacheng Tao, Shijian Lu, and Naoto Yokoya.
\newblock Foundation models for remote sensing and earth observation: A survey.
\newblock {\em arXiv preprint arXiv:2410.16602}, 2024.

\bibitem{xiong2024neural}
Zhitong Xiong, Yi Wang, Fahong Zhang, Adam~J Stewart, Jo{\"e}lle Hanna, Damian Borth, Ioannis Papoutsis, Bertrand Le~Saux, Gustau Camps-Valls, and Xiao~Xiang Zhu.
\newblock Neural plasticity-inspired foundation model for observing the earth crossing modalities.
\newblock {\em arXiv e-prints}, pages arXiv--2403, 2024.

\bibitem{yadav2024unsupervised}
Ritu Yadav, Andrea Nascetti, Hossein Azizpour, and Yifang Ban.
\newblock Unsupervised flood detection on sar time series using variational autoencoder.
\newblock {\em International Journal of Applied Earth Observation and Geoinformation}, 126:103635, 2024.

\bibitem{zheng2021building}
Zhuo Zheng, Yanfei Zhong, Junjue Wang, Ailong Ma, and Liangpei Zhang.
\newblock Building damage assessment for rapid disaster response with a deep object-based semantic change detection framework: From natural disasters to man-made disasters.
\newblock {\em Remote Sensing of Environment}, 265:112636, 2021.

\bibitem{zheng2024towards}
Zhuo Zheng, Yanfei Zhong, Liangpei Zhang, Marshall Burke, David~B Lobell, and Stefano Ermon.
\newblock Towards transferable building damage assessment via unsupervised single-temporal change adaptation.
\newblock {\em Remote Sensing of Environment}, 315:114416, 2024.

\end{thebibliography}
}

\clearpage

\appendix
\section*{Appendix}

\section{Algorithm to Generate Synthetic Dataset}
\label{appendix:alg}
\algrenewcommand\algorithmicrequire{\textbf{Input:}}
\algrenewcommand\algorithmicensure{\textbf{Output:}}

\begin{algorithm}
\caption{Algorithm to Generate Synthetic Dataset}
\begin{algorithmic}[1]
\Require $\mathbf{x}_\text{base}$ and $\mathbf{x}_\text{cut}$ (base image and another image for CutMix from the EuroSAT dataset), $p_1$ and $p_2$ (probabilities to apply seasonal change and cloud cover), $\text{is\_disaster}$ (indicator to apply task-relevant change)
\Ensure $\mathcal{X} = \{\mathbf{x}_1, \dots, \mathbf{x}_5\}$ (generated satellite image time series)
\For{$i \gets 1$ to $4$}
    \Comment{Apply seasonal change and cloud cover to mimic pre-disaster images}
    \State $\mathbf{x}_i \gets \text{ApplyRandomSeasonalChange}(\mathbf{x}_\text{base}, p_1)$
    \State $\mathbf{x}_i \gets \text{ApplyRandomCloudCover}(\mathbf{x}_i, p_2)$
\EndFor
\If{$\text{is\_disaster}$}
    \State $\mathbf{M} \gets \text{GenerateRandomSoftMask}$ 
    \Comment{Generate a random mask with the Gaussian filter for CutMix}
    \State $\mathbf{x}_5 \gets \text{CutMix}(\mathbf{x}_\text{base}, \mathbf{x}_\text{cut}, \mathbf{M})$
\EndIf
\State $\mathbf{x}_5 \gets \text{ApplyRandomSeasonalChange}(\mathbf{x}_5, p_1)$
    \Comment{Only apply seasonal change to the last image to mimic the post-disaster scenario}
\end{algorithmic}
\label{alg:synthetic}
\end{algorithm}

\section{Implementation Details}
\label{appendix:imp}
The original data for the EuroSAT and RaV{\AE}n datasets are based on Sentinel-2, an optical satellite that captures geo-referenced images across 13 spectral bands. For our experiments, we simplify the process by using only the RGB bands (bands 4, 3, and 2), which provide sufficient visual information for detecting extreme events. This choice not only aligns with common practices in optical satellite image analysis but also reduces memory usage during model training.

For the autoencoder setup described in Section \ref{subsec:pipeline}, we employ a weight-shared Vision Transformer \cite{dosovitskiy2020image} as the encoder and a weight-shared transformer-based decoder \cite{carion2020end, he2022masked} to reconstruct the satellite patch time series.
Inspired by the hyperparameter choices in ~\cite{dosovitskiy2020image} and ~\cite{he2022masked}, we set the embedding dimension to 256, patch size to 8, encoder depth to 4, number of heads to 8, and decoder depth to 4, considering our input patch size of 64 $\times$ 64.

\begin{figure}[b]
    \begin{center}
    \includegraphics[width=1\linewidth]{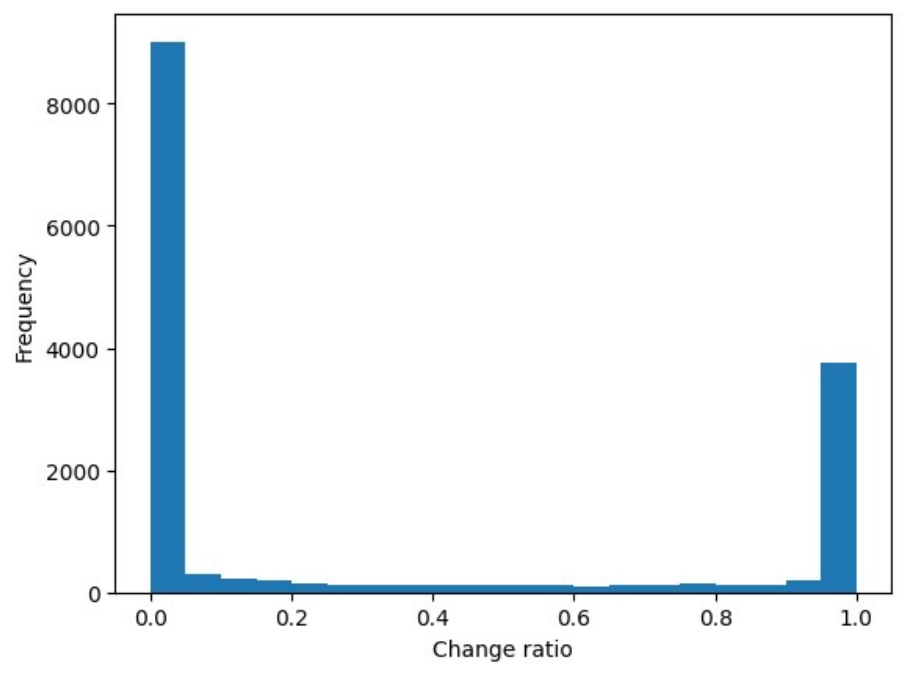}
    \end{center}
    \vspace{-5mm}
    \caption{Histogram of patch \textit{change ratios}, where the \textit{change ratio} is defined as the proportion of changed pixels relative to the total pixels in each patch. The binary distribution shows that most patches are either entirely unchanged (change ratio close to 0) or fully affected by extreme events (change ratio close to 1). This supports the rationale for a patch-level focus in our approach.}
    \label{fig:histogram}
\end{figure}

Our method is implemented using PyTorch~\cite{paszke2019pytorch}. Adam with a weight decay of $1e^{-6}$ is used as the optimizer, and a grid search for learning rates within the range [$1e^{-5}$, $5e^{-5}$, $1e^{-4}$, $5e^{-4}$, $1e^{-3}$, $5e^{-3}$, $1e^{-2}$] revealed that $1e^{-4}$ was the most stable. Cosine annealing with a warmup epoch of 10 serves as the learning rate scheduler. The batch size and learning epoch are uniformly set to 64 and 200, respectively, for all experiments. 
Following standard machine learning protocols, hyperparameter selection is done using the validation set based on the primary evaluation metrics, Average Precision (AP). Based on extensive grid search, we set the loss balancing weights $\lambda=0.5$ and $\mu=0.5$ for the synthetic dataset, and $\lambda=0.25$ and $\mu=0.5$ for the real-world dataset. To complement the primary evaluation metric AP, which is threshold-independent and used for model optimization, we also report the F1 score to demonstrate the model’s performance at a specific threshold. The threshold for the F1 score is selected on the testing set via grid search to identify the value that maximizes F1. It is important to note that this threshold selection is solely for demonstration purposes and does not influence model training or hyperparameter tuning. The primary evaluation remains based on the threshold-independent AP metric to ensure unbiased performance assessment. This procedure is applied consistently to both our method and all baseline methods to ensure fair comparisons. 

All results reported in this paper are based on models trained three times with different random seeds: 42, 43, and 44, respectively. NVIDIA A100 GPUs were used for all experiments.

\begin{figure*}[ht]
    \centering
    \begin{subfigure}[t]{\textwidth}
        \centering
        \includegraphics[width=0.9\textwidth]{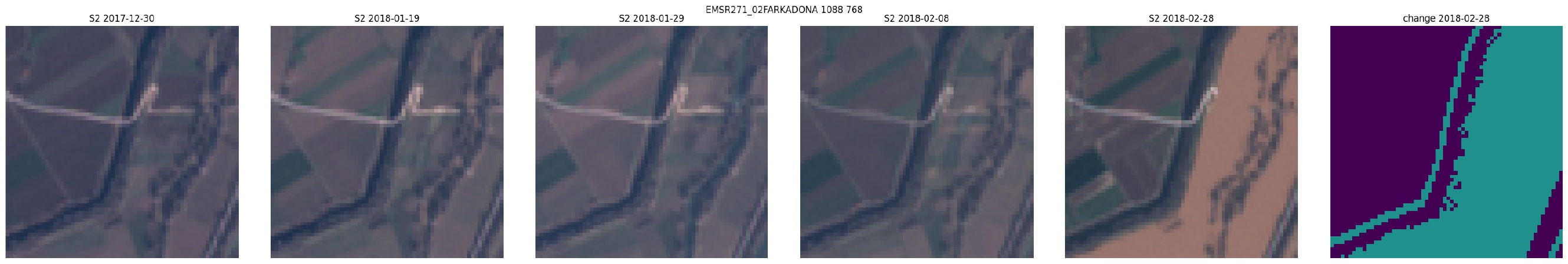}
        \vspace{3mm}
        \includegraphics[width=0.9\textwidth]{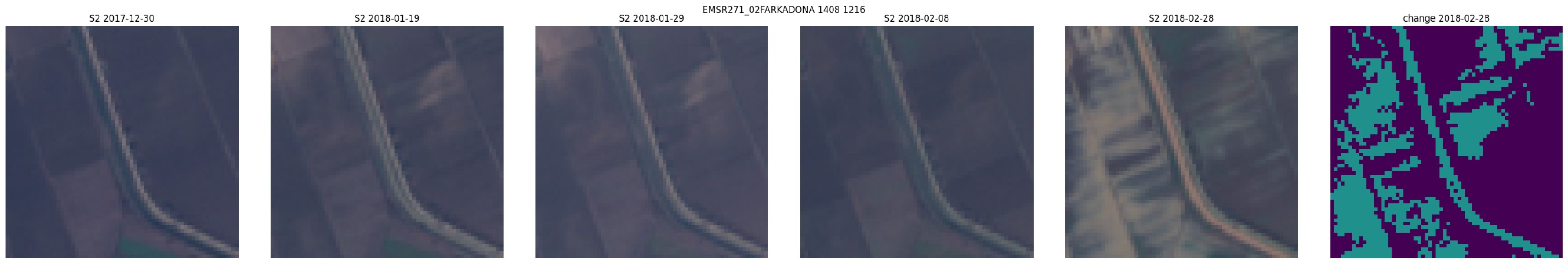}
        \caption{Good annotation examples (flooding): Fine-grained and meaningful pixel-level annotations.}
        \label{fig:good_annotations}
    \end{subfigure}
    
    \vspace{5mm}
    \begin{subfigure}[t]{\textwidth}
        \centering
        \includegraphics[width=0.9\textwidth]{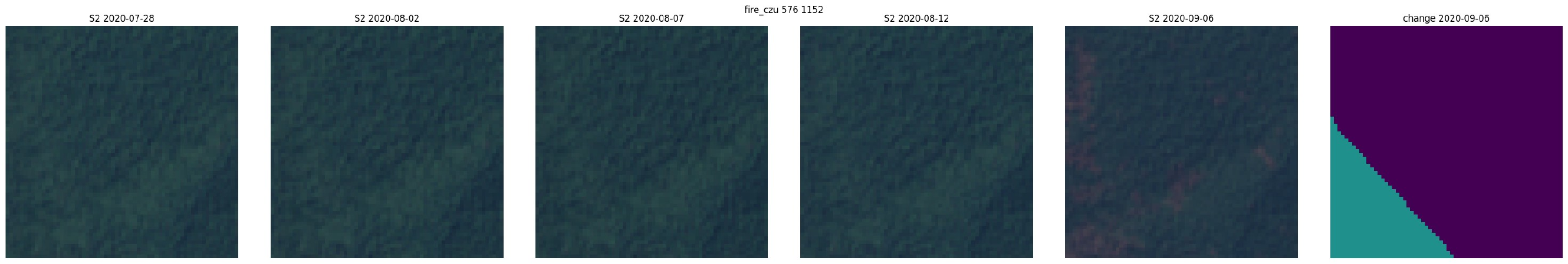}
        \vspace{3mm}
        \includegraphics[width=0.9\textwidth]{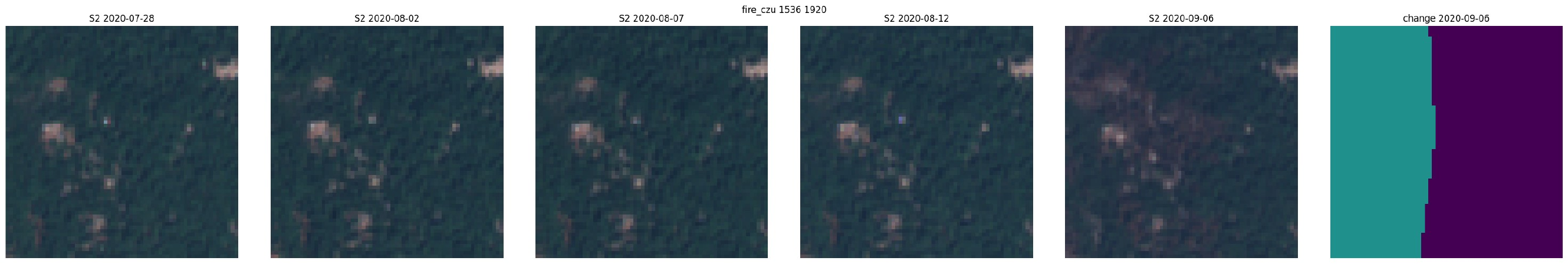}
        \vspace{3mm}
        \includegraphics[width=0.9\textwidth]{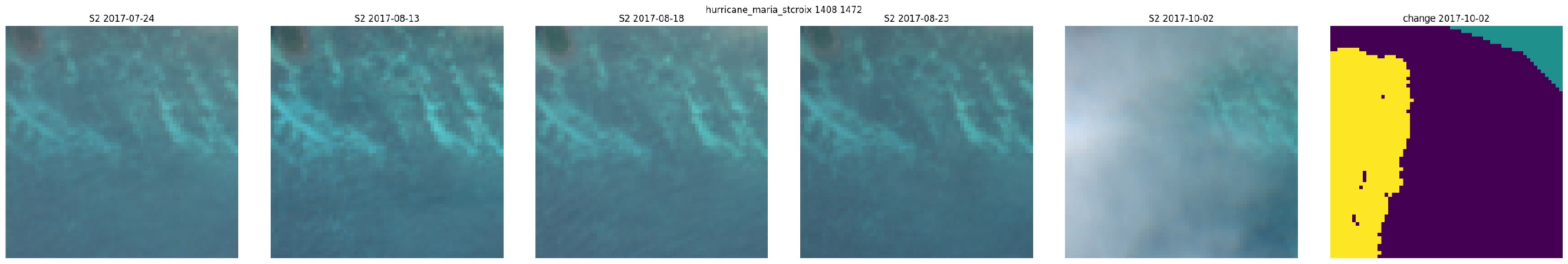}
        \caption{Bad annotation examples: Coarse and unclear pixel-level annotations, making them difficult to interpret.}
        \label{fig:bad_annotations}
    \end{subfigure}
    
    \caption{Examples of pixel-level annotations in the dataset. Good annotations \ref{fig:good_annotations} are meaningful and accurately reflect ground truth, particularly for flooding events, while bad annotations \ref{fig:bad_annotations} are coarse and challenging for model interpretation.}
    \label{fig:annotations}
\end{figure*}

\section{Rationale for Patch-Level Focus: Dataset Characteristics and Challenges}

In this work, we focus on patch-level information rather than pixel-level information for extreme events detection. While pixel-level analysis is a common practice in remote sensing, our choice of patch-level focus stems from specific characteristics of the RaV{\AE}n dataset used in this study. This section highlights the reasoning behind this approach and its implications for future research.

A key characteristic of the RaV{\AE}n dataset is that most patches are either entirely unchanged or entirely affected by extreme events. This binary distribution makes patch-level focus a reasonable choice for effective detection. Figure \ref{fig:histogram} illustrates the histogram of patch statistics, highlighting the dominance of fully-changed and unchanged patches.


Another factor influencing our patch-level approach is the quality of the pixel-level annotations in the dataset. While some annotations, particularly for flooding events, are meaningful and accurately reflect ground truth (e.g., Figure \ref{fig:good_annotations}), many annotations for other disaster types are coarse and often difficult to interpret (e.g., Figure \ref{fig:bad_annotations}), making it challenging for the model to learn from pixel-level annotations, further motivating a patch-level focus.

In summary, the choice of patch-level focus is driven by the dataset’s binary patch characteristics and limitations in pixel-level annotations. Addressing these dataset-specific challenges highlights the need for large-scale multi-disaster datasets with fine-grained, consistent annotations. Such datasets would better reflect real-world disaster response needs, enabling more robust and generalizable models for future research.


\end{document}